# Elephant Search with Deep Learning for Microarray Data Analysis

Mrutyunjaya Panda
P.G. Department of Computer Science and Applications,
Utkal University, Vani Vihar, Odisha-4 India

**1. Introduction**

Gene is the basic unit of storage containing hereditary information in living beings. For the technical point of view,it can be treated as a distinct sequence of nucleotides constituting part of a chromosome. Microarrays data analysis is relatively a new technology that aims to help in finding the right treatment for many diseases with accurate medical diagnosis from the huge number of genes across different experimental conditions. Due to the expensive and complicated nature of the microarray datasets, it is somehow difficult to predict and hence demands careful experimentation with appropriate statistical tools for fruitful analysis. It is well known that the gene expression is a process that maps genes DNA sequence into its corresponding mRNA sequences and then finally to amino acid sequences of proteins. Microarray data analysis is a powerful technology with enormous opportunities presents gene expression profiling to describe the expression levels of hundreds and thousands of genes in cells correlated with corresponding protein, helps one to understand the cellular mechanism of the biological processes. Data Mining helps extract meaningful observations in such a huge and complex microarray gene expressions datasets as a post genome cancer diagnostics to uncover the details on : how the genes are regulated; how genes makes an impact on the cancerous mutation of cells and how the performance depends on various medical experimental conditions to name a few. The microarray dataset presents samples or a condition in rows while the respective genes are provided in a column.

The classification data mining is of great impression of dealing with the patients' gene expression profile for a specific disease in a best possible manner, urges of more research in the area for better predictive accuracy. As the large number of genes are present in the Microarray data analysis, it is always suggested to carry out some potential gene (features) selection algorithms in order to find the most informative genes to reduce the curse of dimensionality. This further may be applied with a best possible classifier to predicting the samples correctly to achieve high accuracy reducing the computational cost and more importantly efficient and effective diagnosis and prognosis can then be customized for the treatment for that patient.

The microarray data analysis needs a clear objective to see its successful implementation for a greater cause of the society at large , as cited by Tjaden and Chen (2006). Clustering is one of the popular technique being used for gene profiling microarray data analysis using K-means clustering and Self organizing maps (SOM) (Sheng-Bo, M.R. and Lok, 2006; Young, 2009). Alshamlan, Badr and Alohali (2013) presents a comprehensive study with objectives and approaches for cancer microarray gene expression analysis and conclude with detailed investigation on the available approached in this area.

Researchers found that most of the cancer studies with microarray gene expression profiling contain comparison of various classes of diseases (Simon, 2009; Wang, Chu, and Xie, 2007) and their predictions, hence sought for using classification algorithms instead of clustering ones (Dougherty, Kohavi, and Sahami, 1995). Support vector machine(SVM) is considered to be one of the most popular and well established classification methods for microarray data analysis for binary classifications initially (Platt, Cristianini, and Shawe-Taylor, 2000). But, as many cancer datasets are of Multi-class, researchers have proposed to use many variants of SVM such as: DAGSVM ((Platt, Cristianini, and Shawe-Taylor, 2000), evolutionary SVM (ESVM) (Huang and F.-L. Chang, 2007), genetic algorithm based SVM (GASVM) (El Akadi, A. Amine, A. El Ouardighi, and D. Aboutajdine, 2009) and Fuzzy SVM (FSVM) (Mao, X. Zhou, D. Pi, Y. Sun, and S. T. C. Wong, 2005).

Neural network based classifiers are also been proposed by many to effective and efficient microarray data analysis that includes: Functional link Neurak network (FNN) (Wang, Chu, and Xie, 2007), Extreme learning machines (ELM) (R. Zhang, G.-B. Huang, N. Sundararajan, and P. Saratchandran, 2007), improved wavelet neural network (WNN) (Zainuddin and P. O, 2009), probabilistic neural networks (PNN) (Berrar, Downes, and Dubitzky, 2003) and subsequent artificial neural network (SAAN) (Roland, Dawn, Holger, Dirk, Klaus, Siegfried, and Mathias, 2004).

Apart from single classification or clustering methods for gene classification, ensemble methods are also been used researchers to Multi-class classification problems, but it is noticed that the ensemble methods are not able to improve the performance in comparison the single classifier methods (Ghorai, A. Mukherjee, S. Sengupta, and P. Dutta, 2010).

The authors (Kothandan and Sumit Biswas, 2016) present a comparison between kernel-based methods and decision trees to explore the best predictive model for identifying miRNAs involved in the cancer pathway.

Considering all the above into consideration, one can conclude that an efficient gene selection method is a must with novel approaches followed by development of a promising fast classifier for better gene prediction with acceptable accuracy.

This motivated us to carry out further experiments on using a novel elephant search based optimization with a deep neural network classifier for improved performance in diverse microarray dataset with two-class, three-class and four class classification.

**Motivation and objective**

Even though there are lots of research opined of using either filter based or wrapper based or a hybrid of these two, for finding a subset of most informative genes for better clinical diagnosis, still there are lot to achieve in terms of performance with new gene selection (feature selection) methods for obtaining new insights in the clinical diagnosis. Considering gene selection is NP hard (Patrenahalli and F. K, 1977) and finding optimal gene from gene expression profiles is really a challenge for getting predictive accuracy. There are several recommendations of using classification and clustering methods to address the problem with adding of novel Multi-objective optimization and some sort of suitable classifier for addressing the binary and Multi-class microarray dataset as a future scope present in the literature. Hence, we are motivated to solve this issue by using an efficient elephant search based optimization with stochastic gradient descent based deep learning classifier. Further, we compared with the approach with already established firefly search optimization for checking the efficacy of our proposed approach. Finally, a comparison with others related work justifies our proposal.

## 2. Material and Methods

This section discusses about the datasets and methodology adopted in this paper.

### 2.1. Datasets used

We use publicly available microarray dataset (Liu, Jinyan Li and Limsoon Wong, 2005; Zhu, Ong and Dash, 2007) for our proposed research in this paper, the details are as provided in Table 1.

**Table 1: Description of the various Microarray Datasets used**

| Sl.No. | Dataset | Details |
|---|---|---|
| 1 | Prostrate Cancer | This dataset contains is used for binary classification to classify the Tumor Vs Normal samples. The training dataset contains 52 prostrate tumor sample and 50 normal samples. This dataset contains 12600 cancer genes. |
| 2 | Leukemia (ALL-AML) | This dataset contains bone marrow samples, collected over 7129 probes from 6817 human genes, out of which 38 samples ( 27 ALL and 11 AML) are for training purposes and 34 samples (20-ALL and 14 AML) for testing purpose. This dataset is used for binary classification. |
| 3 | Colon Tumor | This dataset contains 62 samples out of which 40 samples are negative (tumor biospies are from tumors) and 22 samples are are positive (biospies are from healthy parts of the colons of the same person. Based on the confidence in the measured confidence level, 2000 out of 6500 genes are selected. This fall under binary classification. |
| 4 | DLBCL-Stanford | Diffuse large B-cell lymphoma (DLBCL)dataset contains 47 samples ( 24 from "germinal center B-like" group and 23 are from "activated B-like" group), where each sample is represented by 4026 genes expressions. This is used for 2-class classification. |
| 5 | Lung-H | The Lung-Harvard dataset contains Multi-class (5-class) for classification. This has 203 samples of lung tumors ( 139 sample of lung adenocarcinomas (labeled as ADEN) + 21 of squamous cell lung carcinomas (labelled as SQUA) +20 from pulmonary carcinoids (labeled as COID)+6 from small-cell lung carcinomas (labeled as SCLC) and 17 normal lung samples (labeled as NORMAL) with each sample consisting of 12600 genes. |
| 6 | Ovarian Cancer | The ovarian dataset makes us understand the situation to distinguish the proteomic patterns in serum to have the symptom of ovarian cancer or not. The cancer is mostly significant to the women with similar family history. The dataset is obtained from the proteomic spectra generated by mass spectroscopy with 253 samples (162 ovarian samples and 91 normal ones). The raw spectral data contains the relative intensity amplitude for each sample of 15154 M/Z (molecular mass/charge) identities. This is used for 2-class |

| | | classification. |
|---|---|---|
| 7 | Breast Cancer | This dataset contains samples of 78 patients (34 is relapse category where the distance metastases developed in the patients within 5 years of time and the rest is non-relapse category for healthy patients within the same period of time). The total number of genes present in the dataset is 24481. The value with "NaN" symbol in original ratio data is replaced with 100.0. This is used in binary classification. |
| 8 | MLL | This dataset contains 3-class (conventional acute lymphoblastic (ALL), acute myeloid leukemias (AML) and Mixed-lineage leukemias (MLL)). The MLL translocations are basically found in infant leukemias and chemotheraphy-induced leukemia with a uniform and distinct pattern to classify all the classes. |
| 9 | SRBCT | SRBCT (small round blue cell tumors) dataset contains four different types of childhood cancer tumors such as: Ewing's family of tumors (EWS), neuroblastoma (NB), non-Hodgkin lymphoma (BL) and rhabdomyosarcoma (RMS). The gene expression values of these tumors based on responses to therapy and prognosis with different treatment options with similar appearance of routine histology makes its extremely challenging for classification data mining. |
| 10 | CNS | The CNS (central nervous system) dataset presents the outcome prediction of the patients for embryonal tumor. This contains a total of 60 samples (21 are survivors and 39 are failures) with 7129 number of genes. This is used for 2-class classification problem. |

## 2.2. Methodologies used

This section highlights the gene selection and classification methodology adopted in this paper.

### 2.2.1. Gene selection methods
The gene (or otherwise called as features) selection is of paramount importance for dimension reduction in a huge dataset. The selection of the minimal best genes that represent the original genes in the dataset. This may either lead to give faster result with acceptable accuracy.
Bio-inspired search algorithms are such a popular gene selection methods that seems to address the various complex problems with large scale, NP-hard and multimodal nature; most effectively.
The classical search methods produce local optimum so that they are faster and provides best accuracy in comparison to the global search optimization methods (such as: Genetic algorithm and particle swarm optimization etc.). The local search methods need a good understanding of the initial staring point without which it may not produce effective result, in contrary, the global search methods donot rely on such initial understanding and are less probable in trapping in a local minima (Eslami, Shareef, and Khajehzadeh, 2013).
The following section discussed about the two promising gene selection method for our experimentation.

## A. Firefly search

The firefly search (FFS) was initially coined by Yang (2010) considered to be one among the latest population based global optimization method which works by mimicking the flashing behavior of the fireflies. The simplicity in implementation and efficient computation makes FFS makes it an ideal choice, in comparison to artificial bee colony (ABC), particle Swarm optimization(PSO) and ant colony optimization (ACO) to name a few (Kora and Rama Krishna, 2016). There are around 2000 firefly species which are small insects capable of flashing short and rhythmic light, which in turn attract other other fireflies. Since the light intensity attraction decreases with the distances, the fireflies are only visible up-to several hundreds of meters. The objective function used in the algorithm is linked with the fluorescence light behavior of fireflies. The fireflies move randomly if it found no brighter firefly than it or else it follows the brightest neighboring one.

The advantages of using firefly algorithms lay down with the following points.
- FFA deals with natural way of multimodal optimization by dividing the whole population into subgroups, then each subgroup into local modes and in each local mode, there is existence of global optimum,
- The FFA converge faster through nonlinear attractiveness behaviors among its multiple agents.
- The simplicity makes it popular in use for diverse optimization problems.

## B. Elephant search

The Elephant Search Algorithm (ESA) (Deb, Fong, Tian, 2015) is a highly nonlinear, multimodal global optimization technique inspired by the biological habits of elephant herds, This search algorithm does the following tasks for its implementation.
- The best possible solution is achieved through a iterative process that are represented by the lifetimes of the searching elephants.
- The local searches are led by some chief female elephants to higher likelihood of obtaining best results.
- The male elephants are rangers to lead the elephant clan so that the whole elephant clan go out of local optimum.

While elephant search is being implemented, the following are addressed for their effectivess.
- The visual range of each elephant is fixed and can be calculated using euclidean distance. It can be observed that the visual range is better in male elephants rather than the female ones.
- Secondly, in cases when two or more elephants are searching for visual range, current fitness values are taken for comparison. The elephant with the higher fitness value shall remain and the others can randomly be removed.
- In this basic ESA (elephant search algorithm), it is observed that only a single female elephant group exists and there is no separation of the group.
- In case a elephant die, the new baby elephant of the same sex is born for group's gender balancing and fixed group size.

## 2.3. Deep Learning

Deep learning (Min, Lee, and Yoon, 2016) is termed as universal approximator because of its mapping from input to output as y=f(x) to find out correlation among attributes x and y present in the dataset. Neural networks are modeled based on the working of human brain for pattern recognition. Deep learning (Deep neural network) differs from the conventional neural network in terms of depth, consisting of more than one hidden layers apart from input and output layer. This is why , deep learning is also called as "stacked neural network". Minimum of three hidden layers can be thought of as deep learning. Deep learning further can be seen to have feature hierarchy since they combine and aggregate the features from one layer to the next. This way, it increases complexity and level of abstraction and makes it a good choice for handling the very large and high dimensional complex dataset. The deep neural network needs many hyper-parameter to be set for implementation and at the same time, it is to be noted that finding optimal set of values for those hyper-parameter may not be feasible using gradient descent algorithm due to several constraints like: the dataset is a mix of both real and discrete; each hyper-parameter is difficult to be optimized alone and finding a local minima involves a great deal of time. Initially, the weights of deep neural network is small enough so that the activation function (softmax activation function is used here) operates linearly with large gradient value. The learning rate of the deep neural network should be chosen efficiently so that the validation error is kept at minimum. Further, looking at the input, more network capacity is needed and hence sought for large number of hidden layers. The L1 or L2 regularization scheme is needed to check whether the deep neural network can provide better solutions.

The parameter setting used in this gene selection and classification strategy is shown in Table 2.

**Table 2: Parameter settings in gene selection and Classification**

| Firefly search | Absorption type-0.001 (setting of the absorption coefficient of the firefly population members) |
|---|---|
| | Betamin-0.33 (set the zero distance attractiveness of the firefly members) |
| | Accelerator type- normal |
| | Chaotic coefficient-4.0 |
| | Chaotic Parameter type-Normal |
| | Chaotic population type-Normal |
| | Chaotic mapping type- logistic map |
| | Number of iteartions-20 |
| | Mutation probability-bit flip |
| | Objective type-merits/ mult-iobjective |
| | Population size-20 (particles in the swarm) |
| | Report frequency-20 (set how frequently reports are generated) |

| Elephant search | Accelerator type-Normal |
| --- | --- |
| | Chaotic Coefficient-Normal |
| | Chaotic Parameter type- Normal |
| | Chaotic type- Logistic map |
| | Number of Iterations- 20 |
| | Mutation probability-0.01 |
| | Mutation type-bit flip |
| | Objective type- merits or multi-objective |
| | Population size-20 |
| | Report frequency-20 |
| | Seed-1 |
| Deep learning | Activation function-Softmax |
| | Weight initialization method-XAVIER |
| | Bias initialization-1.0 |
| | Distribution function- Normal Distribution |
| | Learning Rate-0.1 |
| | Bias Learning rate-0.01 |
| | Momentum-0.9 |
| | Updater for stochastic gradient descent- NESTEROVS |
| | Gradient normalization threshold-1.0 |
| | Loss function-Loss MCXENT |
| | ADADELTA's rho parameter-0.0 |
| | ADADELATA epsilon parameter-1.0E-6 |
| | RMSPROP's RMS delay parameter-0.95 |
| | ADAM's Mean decay parameter-0.9 |
| | ADAM's variance parameter-0.999 |
| | Number of Epochs-10 |
| | Optimization algorithm- SGD (stochastic gradient descent) |
| | Batch size-100 |
| | Seed-1 |
| | Number of decimal places-2 |

ADADELTA (Zeiler, 2012) is a per dimension learning rate method basically used for gradient descent method in deep learning classifier, that takes minimum computation overhead, no manual tuning of learning rate (hence dynamic adaptation) and robust to noisy data in selection of hyper-parameters.

ADAM (Kingma and Ba, 2015), is an straightforward and simple stochastic gradient descent optimization method used to adaptive estimate of the lower order moments efficiently. This way, it takes less memory for computation, which is interesting.

## 3. Experimental Findings and Discussion

In this, we present our experimental results and discuss about its effectivess in microarray gene expression profiling. Table 2 and Table 3 demonstrate the proposed approach with FFS and ES based Deep learning respectively.

**Table 3: Firefly search based optimization with Deep learning classifier**

| Sl.No. | Dataset | Original Attibute | Instances | Number of classes | Reduced Attributes | Time in seconds | Accuracy in % |
|---|---|---|---|---|---|---|---|
| 1 | Prostrate Cancer | 12601 | 102 | 2 | 5189 | 1.24 | 87.26 |
| 2 | Leukemia (ALL-AML) | 7130 | 38 | 2 | 2463 | 1.84 | 100 |
| 3 | Colon Tumor | 2001 | 62 | 2 | 562 | 0.43 | 77.43 |
| 4 | DLBCL-Stanford | 4027 | 47 | 2 | 1805 | 0.79 | 89.36 |
| 5 | Lung-Harvard | 12601 | 203 | 5 | 5304 | 15.53 | 93.11 |
| 6 | Ovarian Cancer | 15155 | 253 | 2 | 35 | 17.23 | 97.24 |
| 7 | Breast Cancer | 10 | 286 | 2 | 6 | 0.75 | 65.39 |
| 8 | MLL | 12582 | 72 | 3 | 190 | 14.59 | 80.56 |
| 9 | SRBCT | 2308 | 83 | 4 | 768 | 0.63 | 93.98 |
| 10 | CNS | 7129 | 72 | 2 | 1526 | 0.45 | 56.67 |

**Table 4: Elephant search based optimization with Deep learning classifier**

| Sl.No. | Dataset | Original Attributes | Instances | Number of classes | Reduced Attributes | Time in seconds | Accuracy in % |
|---|---|---|---|---|---|---|---|
| 1 | Prostrate Cancer | 12601 | 102 | 2 | 4267 | 1.33 | 88.24 |
| 2 | Leukemia (ALL-AML) | 7130 | 38 | 2 | 1044 | 0.31 | 92.11 |
| 3 | Colon Tumor | 2001 | 62 | 2 | 572 | 0.41 | 79.03 |
| 4 | DLBCL-Stanford | 4027 | 47 | 2 | 1717 | 0.44 | 91.49 |
| 5 | Lung-Harvard | 12601 | 203 | 5 | 4545 | 2.92 | 94.10 |
| 6 | Ovarian Cancer | 15155 | 253 | 2 | 384 | 1.67 | 99.21 |
| 7 | Breast Cancer | 10 | 286 | 2 | 6 | 1.09 | 73.43 |
| 8 | MLL | 12582 | 72 | 3 | 190 | 15.13 | 80.56 |
| 9 | SRBCT | 2308 | 83 | 4 | 306 | 0.66 | 83.14 |
| 10 | Central Nervous System (CNS) | 7129 | 72 | 2 | 1621 | 0.47 | 53.34 |

It can be observed from table 3 and Table 4 that Deep learning works well for almost in all datasets except for CNS dataset. This may be due to the less number of instances available for classification.

Further, a comparative study is provided in Table 5, for the understanding the efficacy of the proposed approach. The comparison opines that the proposed approach achieves comparable accuracy for all the microarray datasets. While verifying the suitability with others work as given in Table 5, it is observed that Vural and Subasi (2015) have used singular value decomposition along with information gain to reduce the number of attributes till they obtain less than that of the number of samples and Mukkamala et al. (2005) presented their work with different number of reduced attributes, more number of attributes less is the accuracy for them.

**Table 5: Accuracy (%) Comparison with some existing research**

| Method/Dataset | ALL-AML | Colon Tumor | SRBCT | Lung-H | DLBCL | Prostrate Cancer |
|---|---|---|---|---|---|---|
| SVM Vural and Subasi (2015) | 97.14 | 83.87 | 95.18 | 93.6 | 98.7 | --- |
| ANN Vural and Subasi (2015) | 91.43 | 83.87 | 95.18 | 92.12 | 94.81 | ---- |
| Random Forest Vural and Subasi (2015) | 91.43 | 87.1 | 86.75 | 90.64 | 90.91 | ---- |
| PSA (Glinsky et al., 2004) | --- | --- | ---- | --- | --- | 77 |
| MARS with 6 attributes Mukkamala et al. (2005) | --- | ---- | ---- | ---- | ----- | 68.2 |
| Random forest with 6 attributes Mukkamala et al. (2005) | --- | --- | ---- | --- | --- | 80.2 |
| LGP with 6 attributes Mukkamala et al. (2005) | --- | ---- | ---- | ---- | ----- | 92.1 |
| Ours (FFS+DL) | 100 | 77.42 | 93.98 | 93.11 | 89.36 | 87.26 |
| Ours (ES+DL) | 92.11 | 79.03 | 83.14 | 94.1 | 91.49 | 88.24 |

**Discussions**

Deep learning fails to perform better since it may over-fit when a complicated model is chosen to learn from an easy problem. Also, the Deep learning depends largely on the depth of the network, it is envisaged of working well in complex problems. It is well understood that direct Multi-class classification results low accuracy in comparison to two class classification. At the same time, its a very challenging job to obtain high accuracy in the microarray dataset because of less number of samples in comparison number of attributes even after reduction.

**4. Conclusion and Future scope**

From the literature search and experiments conducted in this paper, it is understood that a reliable and detailed analysis is most essential in the success of the gene expression data analysis for cancer classification. As the gene expression datasets are complex ones, novel gene selection followed by an effective and efficient classification strategy needs utmost care for diagnosis of the disease. The result obtained opined that elephant search method could select most appropriate genes out many redundant genes present in the dataset. The most promising , very recent deep learning classification technique is then found a promising ones with good accuracy. Further, the effectiveness of our proposed method is to be tested in big dataset with large number of samples along with a large number of attributes in future.

# References

B. Tjaden and J. Cohen, A survey of computational methods used in microarray data interpretation, Applied Mycology and Biotechnology, Bioinformatics, 2006; 6: 7–18.

G. Sheng-Bo, L. M. R., and T.-M. Lok, "Gene selection based on mutual information for the classification of multi-class cancer," in Proceedings of the 2006 international conference on Computational Intelligence and Bioinformatics - Volume Part III, ser. ICIC'06. Springer-Verlag, 2006, pp. 454–463.

Y. T. Young, "Efficient multi-class cancer diagnosis algorithm, using a global similarity pattern," Comput. Stat. Data Anal., vol. 53, no. 3, pp. 756–765, Jan. 2009. [Online]. Available: http://dx.doi.org/10.1016/j.csda.2008.08.028

Hala M. Alshamlan, Ghada H. Badr, and Yousef Alohali, A Study of Cancer Microarray Gene Expression Profile: Objectives and Approaches, Proceedings of the World Congress on Engineering 2013 Vol II, WCE 2013, July 3 - 5, 2013, London, U.K.,

R. Simon, "Analysis of dna microarray expression data," Best practice and research Clinical haematology, vol. 22, no. 2, pp. 271–282, 2009.

L. Wang, F. Chu, and W. Xie, "Accurate cancer classification using expressions of very few genes," IEEE/ACM Transactions on Computational Biology and Bioinformatics, vol. 4, no. 1, pp. 40–53, 2007.

J. Dougherty, R. Kohavi, and M. Sahami, "Supervised and unsupervised Discretization of continuous features," in machine learning: Proceedings of the twelfth international conference. Morgan Kaufmann, 1995, pp. 194–202.

J. C. Platt, N. Cristianini, and J. Shawe-taylor, "Large margin dags for multiclass classification," in Advances in Neural Information Processing Systems. MIT Press, 2000, pp. 547–553.

H.-L. Huang and F.-L. Chang, "ESVM: Evolutionary support vector machine for automatic feature selection and classification of microarray data," Biosystems, vol. 90, no. 2, pp. 516 – 528, 2007.

A. El Akadi, A. Amine, A. El Ouardighi, and D. Aboutajdine, "A new gene selection approach based on minimum redundancy-maximum relevance (mrmr) and genetic algorithm (ga)," in Computer Systems and Applications, 2009. AICCSA 2009. IEEE/ACS International Conference on, 2009, pp. 69–75.

Y. Mao, X. Zhou, D. Pi, Y. Sun, and S. T. C. Wong, "Multiclass cancer classification by using fuzzy support vector machine and binary decision tree with gene selection," Journal of Biomedicine and Biotechnology, vol. 2, no. 8, pp. 160–171, 2005.

R. Zhang, G.-B. Huang, N. Sundararajan, and P. Saratchandran, "Multicategory classification using an extreme learning machine for microarray gene expression cancer diagnosis," Computational Biology and Bioinformatics, IEEE/ACM Transactions on, vol. 4, no. 3, pp. 485–495, 2007.

Z. Zainuddin and P. O, "Improved wavelet neural network for early diagnosis of cancer patients using microarray gene expression data," in Neural Networks, 2009. IJCNN 2009. International Joint Conference on, 2009, pp. 3485–3492.

D. Berrar, S. Downes, and W. Dubitzky, "Multiclass cancer classification using gene expression profiling and probabilistic neural networks," in Pacific Symposium on Biocomputing, vol. 8, 2003, pp. 5–16.

L. Roland, D. Dawn, S. Holger, T. Dirk, R. Klaus, P. Siegfried, and W. Mathias, "The subsequent artificial neural network (sann) approach might bring more classificatory power to ann-based dna microarray analyses," Bioinformatics, vol. 20, no. 18, pp. 3544–3552, 2004.

S. Ghorai, A. Mukherjee, S. Sengupta, and P. Dutta, "Multicategory cancer classification from gene expression data by multiclass nppc ensemble," in Systems in Medicine and Biology (ICSMB), 2010 International Conference on, 2010, pp. 4–48.

Ram Kothandan and Sumit Biswas, Comparison of Kernel and Decision Tree-Based Algorithms for Prediction of MicroRNAs Associated with Cancer, *Current Bioinformatics*, 2016, *11*, 143-151.

N. Patrenahalli and F. K, "A branch and bound algorithm for feature subset selection," Computers, IEEE Transactions on, vol. 26, no. 9, pp. 917–922, 1977.

Huiqing Liu, Jinyan Li and Limsoon Wong, Use of extreme patient samples for outcome prediction from gene expression data, Bioinformatics, Vol. 21 no. 16 2005, pp. 3377–3384. Oxford University Press.

Zexuan Zhu, Y. S. Ong and M. Dash, Markov blanket- Embeded genetic Algorithm for gene selection, Pattern recognition, Vol. 49, No. 11, pp. 3236-3248, 2007.

Mahdiyeh Eslami, Hussain Shareef, and Mohammad Khajehzadeh, Firefly Algorithm and Pattern Search Hybridized for Global Optimization, in: D.-S. Huang et al. (Eds.): ICIC 2013, LNAI 7996, pp. 172–178, 2013., Springer.

Yang, X.-S.: Firefly Algorithm, Stochastic Test Functions and Design Optimisation. Int. J. Bio-Inspired Computation 2(2), 78–84 (2010)

Padmavathi Kora and K. Sri Rama Krishna, Hybrid firefly and Particle Swarm Optimization algorithm for the detection of Bundle Branch Block, International Journal of the Cardiovascular Academy 2 (2016) 44–48, Elsevier

Suash Deb, Simon Fong, Zhonghuan Tian, Elephant Search Algorithm for Optimization Problems, The Tenth International Conference on Digital Information Management (ICDIM 2015), 249-255, IEEE.

Seonwoo Min, Byunghan Lee, and Sungroh Yoon, Deep Learning in Bioinformatics, https://arxiv.org/ftp/arxiv/papers/1603/1603.06430.pdf

Matthew D. Zeiler, ADADELTA: An adaptive learning rate method, arXiv:1212.5701v1 [cs.LG] 22 Dec 2012

Diederik P. Kingma and Jimmy Lei Ba, ADAM: A method for stochastic optimization, arXiv:1412.6980v8 [cs.LG] 23 Jul 2015.

H. Vural and A. Subasi, Data Mining techniques to classify Microarray gene expression data using selection by SVD and information gain, Modeling of Artificial inteligence, vol. 6, no-2 pp. 171-182, 2015.

G. V. Glinsky, A. B. Glinskii, A. J. Stephenson, R. M. Hoffman and W. L. gerald, gene expression profiling predicts clinical outcomes of prostare cancer, The journal of clinical investigation, vol. 113, no.6, pp. 913-923, 2004.

S. Mukkamala, Q. Liu, R. Veeraghattam and A. H. Sung, Computational Intelligent techniques for tumor classifications(using microarray gene expression data), International Journal of lateral computing, vol. 2, no.1, pp. 38-45, 2005.